\documentclass[10pt,twocolumn,letterpaper]{article}

\usepackage[pagenumbers]{cvpr} 
\usepackage{caption}
\usepackage{enumitem}
\usepackage{amsmath}
\usepackage{makecell}
\usepackage{adjustbox}

\definecolor{cvprblue}{rgb}{0.21,0.49,0.74}
\usepackage[pagebackref,breaklinks,colorlinks,allcolors=cvprblue]{hyperref}

\def\paperID{*****} 
\def\confName{CVPR}
\def\confYear{2026}

\title{StateFlow: Building, Evolving, and Accessing 3D World States for Previsualization}

\author{
\begin{tabular}{c}
Yuyang Yin$^{1,}$\thanks{Equal contribution. $^\dagger$Corresponding author.}
\quad
Zixiang Li$^{1,*}$
\quad
Longxuan Deng$^{1,*}$
\quad
Hongkai Li$^{1}$
\quad
Shifang Zhao$^{1}$
\\
Junnan Liu$^{1}$
\quad
Weirong Huang$^{1}$
\quad
Mengyu Wang$^{1}$
\quad
Tianxiao Fu$^{2}$
\quad
Yikai Wang$^{3}$
\quad
\\
Peng-Shuai Wang$^{4}$
\quad
Xiaojie Jin$^{1}$
\quad
Yao Zhao$^{1}$
\quad
Yunchao Wei$^{1,5,\dagger}$
\\[1.2ex]
$^{1}$Beijing Jiaotong University
\quad
$^{2}$Mootion AI
\quad
$^{3}$Beijing Normal University
\\
$^{4}$Peking University
\quad
$^{5}$Beijing Academy of Artificial Intelligence
\\[1.2ex]
\href{https://yuyangyin.github.io/StateFlow/}{https://yuyangyin.github.io/StateFlow/}
\end{tabular}
}

\makeatletter
\let\stateflowOriginalMakeTitle\@maketitle
\def\@maketitle{%
  \stateflowOriginalMakeTitle
  \begin{center}
    \includegraphics[width=1.0\textwidth]{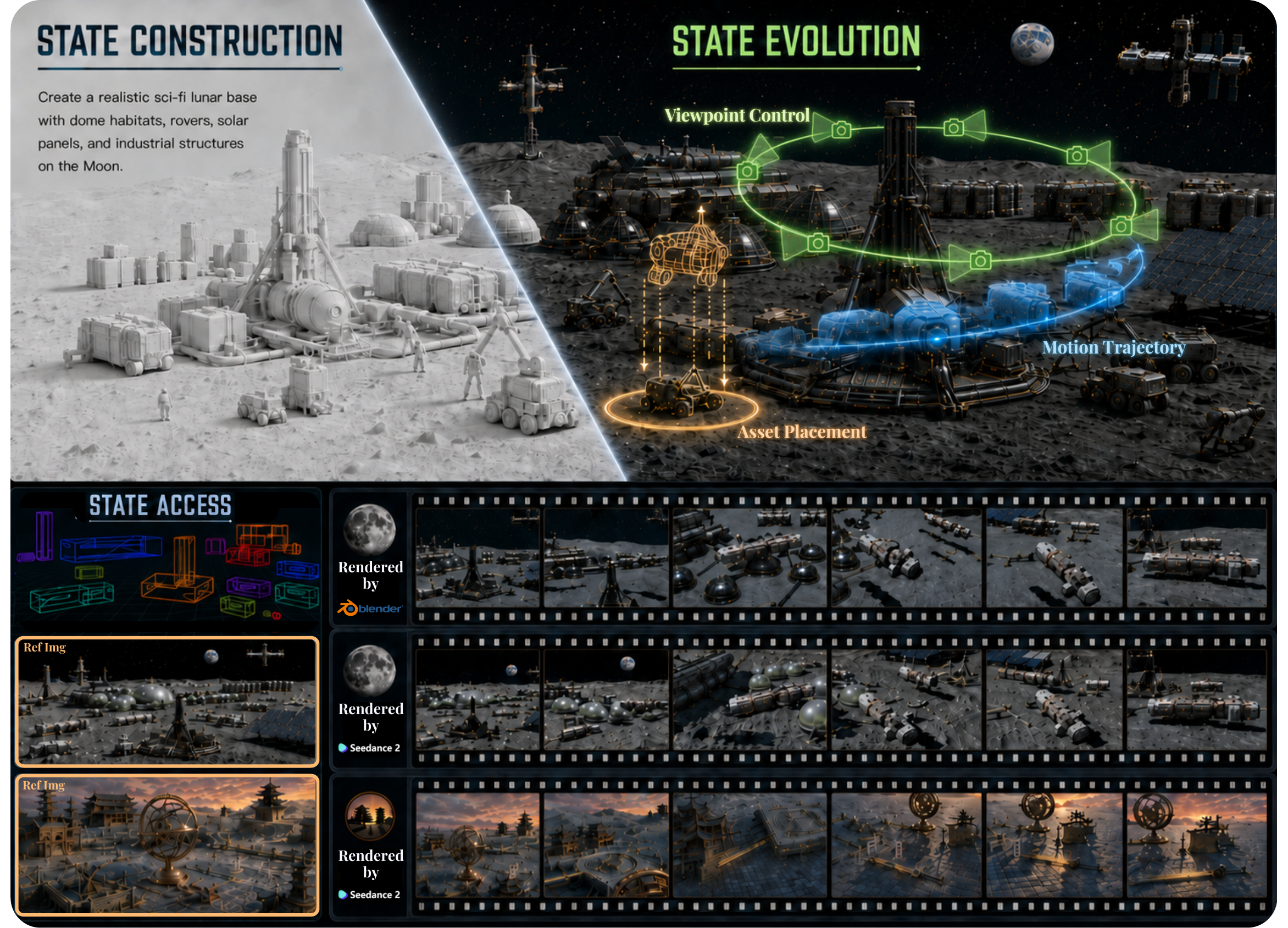}
    \vspace{-0.5em}
    \captionof{figure}{StateFlow turns a creative prompt into a complete, editable 3D world.
    Creators can construct scenes from reference images, expand environments, change visual styles, control object motion, stage dynamic events, and direct cinematic cameras through high-level intent.
    The same persistent world can be reused for cinematic video creation, controllable shot planning, and game-like interactive prototyping.}
    \label{fig:teaser}
  \end{center}%
}
\makeatother

\begin{document}
\maketitle

\begin{abstract}

Previsualization is an intermediate layer between ideas and production in film, games, architecture, and urban design. It lets creators iteratively refine scenes, actions, cameras, and spatial-temporal dynamics. Yet existing generative methods rely on simple prompts to jointly control all of these factors through one-shot image or video synthesis, offering weak controllability and limited support for iterative editing. Fundamentally, a world comprises multiple elements with geometry, appearance, and other attributes, together with cameras. Different frames are produced through local modifications or recombinations of this shared state, which is otherwise largely reused. Therefore, we argue that the missing component is an explicit and persistent working state.
To address this, we present \textbf{StateFlow}, a state-centric framework for generative previsualization. \textbf{Rather than generating videos in one shot, StateFlow uses an editable 3D world to organize scene structure, evolution, and cameras, while off-the-shelf video models enhance visual quality when higher fidelity is desired.} This world is maintained as a persistent structured 3D state of scene elements and camera configurations, serving as the core working representation for previsualization.
Built on this insight, StateFlow has three stages to construct, evolve, and access the world state. State construction lifts generated 2D content into a coherent 3D world through prior-guided, conflict-aware dual-view initialization, while State evolution translates user intent into structured state transitions while preserving world memory, avoiding full-scene regeneration for each edit. State access uses render-feedback reflection to refine camera plans into visually feasible trajectories, avoiding reliance on VLM semantics alone. Experiments show that StateFlow produces high-quality 3D worlds for video creation and game-like prototyping.

\end{abstract}

\section{Introduction}

Previsualization is a fundamental step in filmmaking, game development, and architectural or urban design, where creators plan scenes, block actions, test cameras, and explore spatiotemporal dynamics before final production~\cite{ardal2019collaborative,christie2008camera,halperin2025camera}. Unlike final-content generation, previsualization prioritizes communicating creative intent and enabling rapid iteration over producing polished visual outputs. Creators repeatedly refine scene layout, object placement, motion, and viewpoints to explore different design choices and narrative possibilities~\cite{lino2011director}.

In practice, building previsualization content remains time-consuming and expertise-intensive. Meanwhile, existing generative approaches are poorly suited to this iterative process. Although recent image and video generation methods have substantially improved visual quality~\cite{rombach2022high,brooks2024video,yang2024cogvideox,wan2025wan,bruce2024genie}, they are largely designed for one-shot synthesis: a simple prompt must jointly specify scene content, layout, motion, and cameras, while the generated observations expose no persistent state for local editing and validation. Consequently, they often exhibit spatiotemporal inconsistency, identity drift, and unstable local details when content is modified, extended, or viewed from different cameras.

Fundamentally, a previsualization world consists of multiple scene elements with geometry, spatial poses, semantic and appearance attributes, and dynamic states, together with camera configurations. Different shots largely reuse this information, modifying or recombining only a subset of the shared state. This suggests that previsualization should be modeled through an explicit and persistent working state rather than a sequence of isolated visual outputs.

An explicit 3D state is essential for previsualization: a shared coordinate system preserves object identity and spatial relations, supports local edits to placement and motion, and provides geometric constraints for camera planning across viewpoints and time. Existing methods can already assemble retrieved or generated assets into coherent multi-object 3D scenes~\cite{yang2024holodeck,feng2023layoutgpt,wang2025worldgen,shi2025scenemaker,engstler2025syncity}. However, they commonly treat scene construction as the endpoint, producing a largely static world without jointly modeling its subsequent evolution or repeated, camera-controlled access. For previsualization, an initial scene is only a starting point: the same world must remain editable as it changes and as it is viewed from different cameras.

Building on this formulation, we propose \textbf{StateFlow}, a state-centric framework for generative previsualization. Rather than generating videos in one shot, StateFlow uses an editable 3D world to organize scene structure, evolution, and cameras. Each world is represented by an editable \textbf{\textit{structured 3D state}} of scene elements, geometry, spatial poses, and semantic states; StateFlow constructs and updates this state from user intent, while controllable camera trajectories provide access for scene inspection and downstream previsualization.
When greater visual fidelity is desired, off-the-shelf video models~\cite{seedance2026seedance} enhance geometry-grounded observations into expressive videos without altering the underlying world.

 As shown in Fig.~\ref{fig:teaser}, our framework addresses three core questions in previsualization: how to initialize the structured 3D world state through \textbf{State Construction}, how to evolve the world state according to the real physical world and user intent through \textbf{State Evolution}, and how to use it for downstream tasks through \textbf{State Access}.

In \textbf{State Construction}, StateFlow translates a user's natural-language intent into an initial 3D scene by first generating complementary visual references with an image model. Since a single view cannot reliably capture both object detail and global layout, we introduce {\textit{Prior-Guided Conflict-Aware Dual-View Initialization}}, which combines front-view semantic grounding with BEV spatial grounding. The framework detects and resolves cross-view conflicts in object counts and spatial hypotheses, producing plausible 3D placements.

In \textbf{State Evolution}, StateFlow introduces {\textit{Intent-Guided Structured State Transition}}, which translates user intent into compact updates over the structured state table, rather than continuously simulating dynamics or regenerating videos. It supports scene expansion, local style changes, object pose and motion edits, and discrete asset substitutions for events such as destruction. This keeps world evolution editable, memory-aware, and reusable for later access.

In \textbf{State Access}, StateFlow introduces \textit{World-State Camera Planning with Render-Feedback Reflection}, a two-system design that couples VLM-based camera proposal with render-feedback verification. The VLM proposes trajectories from director intent and shot style, while geometry-faithful rendering exposes visibility, occlusion, framing, and motion issues to guide local repairs. This yields visually grounded trajectories without training a camera policy.

Overall, our formulation shifts previsualization from generating isolated visual outputs to constructing editable, evolvable, and interactive 3D worlds. Operating on the same persistent 3D world, our framework naturally supports two representative downstream previsualization applications: \textbf{\textit{video creation}}, including storyboard, shot planning, and video production, and \textbf{\textit{3D game prototyping}} for interactive content design. Extensive experiments across diverse tasks and application settings show that the 3D worlds constructed by our framework serve as effective representations for downstream applications, demonstrating strong flexibility, controllability, and extensibility.

{In summary, our main contributions are as follows:}
\begin{enumerate}[leftmargin=*]
\item  We introduce a new formulation of generative previsualization as persistent 3D world-state modeling, shifting the goal from one-shot visual synthesis to building editable, evolvable, and reusable 3D worlds.

\item We propose StateFlow, a unified state-centric framework that constructs, evolves, and accesses 3D world states through prior-guided initialization, intent-guided state transition, and render-feedback camera planning.

\item Experiments show that our method produces high-quality and fine-grained results, while naturally supporting two representative previsualization applications: video creation for storyboard, shot planning, and video production, and 3D game prototyping for interactive content design.
\end{enumerate}

\section{Related Work}
\subsection{Previsualization}
Previsualization has been widely explored across film, games, and design as an intermediate process for validating scene layout, camera motion, staging, and collaboration before full production. Early studies emphasized game engines and real-time 3D tools for shot planning and preliminary scene prototyping in film and game pipelines \cite{nitsche2008experiments,northam2012collaborative}. Later work extended previsualization to immersive and collaborative settings, including virtual reality, tangible interfaces, and video-based rapid ideation \cite{ardal2019collaborative,fei2023tape,jo2024collagevis}. Beyond filmmaking, related studies in HCI and architectural design have framed previs as a broader method for design futuring, immersive review, and stakeholder communication \cite{ivanov2022one,yu2022systematic}.

\subsection{3D Scene Generation}

Recent 3D, 4D, and video-based generation methods create detailed objects, dynamic content, and navigable scenes from images, text, or video~\cite{tang2024dreamgaussian,zhao2025hunyuan3d,xiang2025structured,sam3dteam2025sam3d3dfyimages,lin2025partcrafter,ren2023dreamgaussian4d,yin20234dgen,liang2024diffusion4d,xing2025tip4gen,sun2024dimensionx,yang2025matrix,yin2025panoworld,team2026hy,li2026martian,sun2025worldplay,zhang2026worldstereo}. However, they primarily generate individual assets, short dynamic sequences, or video/world reconstructions, without maintaining the persistent spatial relations, semantics, motion, and camera access required for an editable multi-object world. StateFlow instead treats video as an observation of such an explicit world state.

Existing 3D scene generation methods mainly follow three directions. Procedural approaches create large environments with hand-designed rules or asset placement pipelines~\cite{infinigen2023infinite, infinigen2024indoors}, but their diversity and realism are constrained by predefined rules and assets. LLM and VLM based methods infer object layouts and instantiate retrieved assets from curated libraries~\cite{yang2024holodeck,yang2024llplace,sun2025layoutvlm,feng2023layoutgpt,gu2025artiscene,zhu2025imaginarium}; while controllable, they are often limited by the coverage and style of the asset database. Recent 3D generative methods synthesize scene elements or full scenes more flexibly~\cite{sam3dteam2025sam3d3dfyimages,wang2025worldgen,shi2025scenemaker,lin2025partcrafter,huang2025midi,yang2026flowscene,yang2025mmgdreamer,engstler2025syncity}, but accurate placement, global consistency, and high-detail scene quality remain challenging, especially under limited 3D scene data and domain coverage. In contrast, our method organizes generated content into an explicit editable 3D world state, aiming to support open-ended previsualization across diverse styles without relying on fixed scene datasets or asset domains.                                                                                                                                                                                                                                                                                                                                                                  
\subsection{Agentic Video Generation}
Recent video-generation backbones, including Stable Video Diffusion (SVD)~\cite{blattmann2023stable}, CogVideoX~\cite{yang2024cogvideox}, HunyuanVideo~\cite{kong2024hunyuanvideo}, Seedance~\cite{seedance2026seedance}, and Wan~\cite{wan2025wan}, provide increasingly capable primitives for short-clip synthesis from text or images. Building on these models, MovieAgent~\cite{wu2025automated}, AniMaker~\cite{shi2025animaker}, and AniME~\cite{zhang2025anime} use hierarchical planning, candidate-clip selection, and director-centered coordination to generate longer, story-driven videos; VideoClaw~\cite{hitsztmg2026videoclaw}, Toonflow~\cite{hbai2026toonflow}, and ViMax~\cite{huang2026vimax} further expose editable production stages and coordinate assets, shots, and consistency across multiple scenes. Unlike systems centered on scripts, storyboards, or dependencies, StateFlow maintains a unified 3D state of objects, relations, motion, and cameras, using rendered video as feedback for state and camera refinement.

\section{Method}
\begin{figure*}
    \begin{center}\vspace{-0.2cm}
        \includegraphics[width=0.98\linewidth]
        {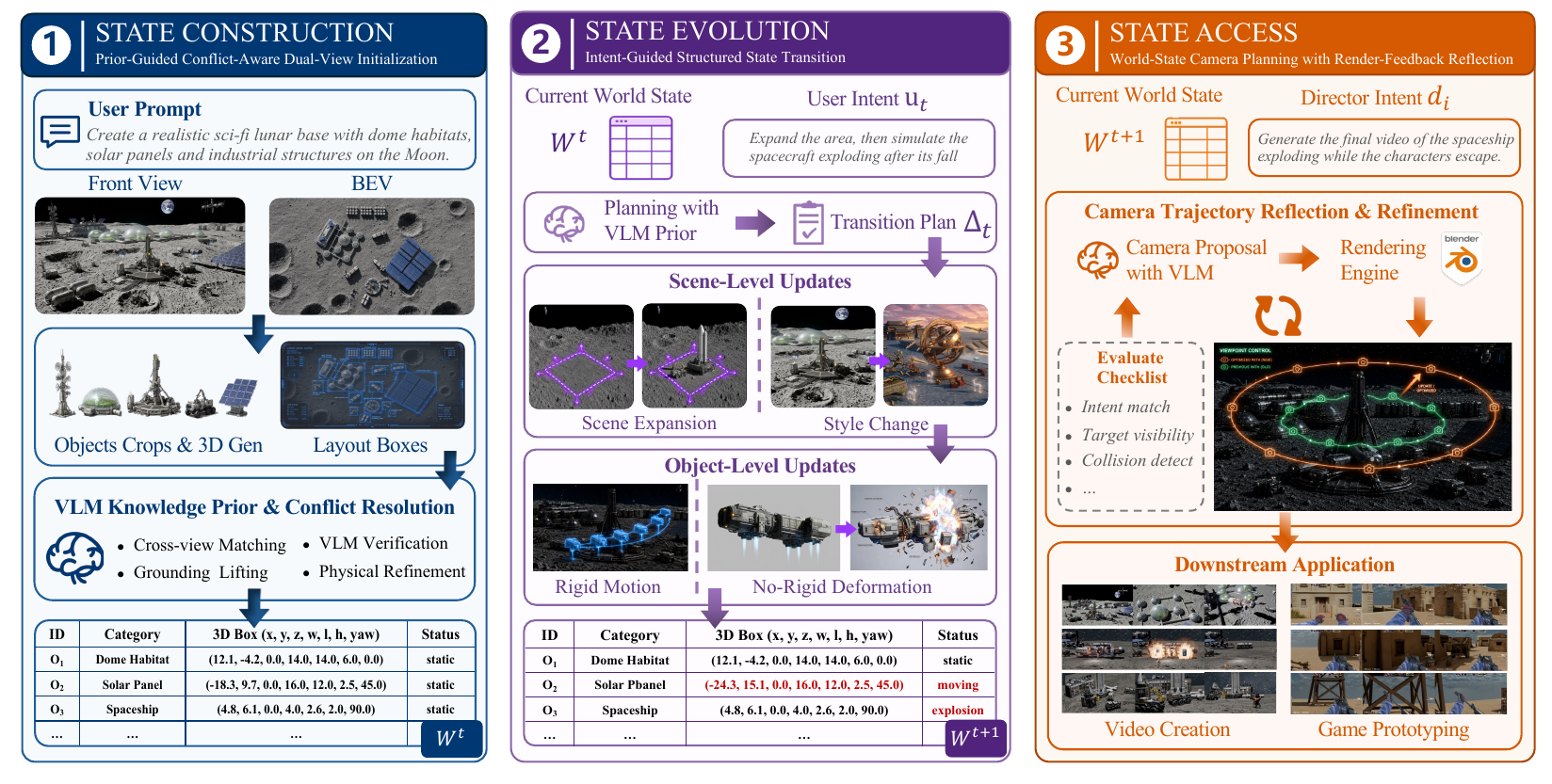}
        \vspace{-1mm}
        \caption{StateFlow turns creative intent into a persistent 3D world that can be constructed, evolved, and accessed for downstream creation.
State Construction resolves cross-view conflicts between appearance, layout, and physical plausibility to build coherent object-centric worlds.
State Evolution applies intent-guided world updates for scene expansion, style changes, object motion, and event transitions while preserving continuity.
State Access refines camera plans with rendered feedback to ensure visibility, framing, and motion feasibility.
The same world state serves both cinematic video creation and game prototyping.}
        \label{fig:method}
    \end{center}
    \vspace{-0.15cm}
\end{figure*}

We present \textbf{StateFlow}, a state-centric framework for previsualization. The key idea is to treat previsualization not as one-shot output generation, but as maintaining a persistent world state that can be constructed, evolved, and accessed under different controls. In the following, we first formulate this process as persistent 3D world-state modeling, then describe how StateFlow instantiates it with an object-centric structured 3D state.

\subsection{Previsualization Formulation}
Both manual previs and generative previs pursue the same process: creators first establish a scene, then modify how it evolves, and finally inspect it through cameras or interactions. The difference is that manual workflows naturally maintain an explicit working state, while current generative methods often treat this state into isolated images or videos. From this perspective, previsualization can be summarized as a state-based process. Given input conditions $\mathcal{C}$, a system first builds an initial world state, then updates this state over time, and finally produces observations under controllable access:
\begin{equation}
\mathcal{W}^0 = \mathcal{F}_{\mathrm{build}}(\mathcal{C}), \quad
\mathcal{W}^{t+1} = \mathcal{F}_{\mathrm{evolve}}(\mathcal{W}^t), \quad
y^t = \mathcal{F}_{\mathrm{access}}(\mathcal{W}^t).
\end{equation}
where $\mathcal{W}^t$ denotes the world state at time $t$, and $y^t$ is the resulting observation. 
This formulation separates three roles that are often entangled in generative visual models: the world state ${W}^t$, its evolution $\mathcal{F}_{\mathrm{evolve}}$, and the rendered or interactive observation $y^t$. In previsualization, this separation is crucial because creators repeatedly modify the same world and inspect it from different viewpoints, rather than consuming a single generated output.

\subsection{Overall Pipeline}
StateFlow instantiates $\mathcal{W}^t$ as an object-centric \textbf{structured 3D state}. We use an object-level state because previsualization needs to reason about what entities exist, where they are, how they look, and how they can change. At time $t$, the world is represented as:
\begin{equation}
\mathcal{W}^t = \{ o_i^t \}_{i=1}^{N_t}, \quad o_i^t = (g_i^t, p_i^t, s_i^t).
\end{equation}
Here, $o_i^t$ denotes the $i$-th object entity in the world state at time $t$, represented by its geometry, spatial placement, and semantic attributes. This is a minimal complete state for our setting: $g_i^t$ provides the geometry for rendering and physical reasoning, $p_i^t$ defines the object's 3D position and pose for spatial layout, and $s_i^t$ stores semantic and editable attributes such as category, appearance, and state descriptors. Removing any of these would respectively lose visual realization, spatial consistency, or controllable semantic evolution.

Built on this representation, StateFlow contains three stages: \emph{State construction} initializes $\mathcal{W}^0$ from input conditions, \emph{state evolution} updates the world over time, and \emph{state access} renders or interacts with $\mathcal{W}^t$ under camera and user controls. The overall pipeline is shown in Fig.~\ref{fig:method} This unified state representation enables automatic generation, local modification, temporal evolution and repeated observation within the same 3D world.

\subsection{State Construction}
Given input conditions $\mathcal{C}$, the scene construction stage initializes the first world state $\mathcal{W}^0$ by estimating both object geometry and global spatial layout. Existing generation pipelines often rely on a single view for scene construction, which either preserves appearance without reliable layout or provides layout without sufficient object appearance and vertical structure. To combine these complementary cues, we first generate a front-view image for object assets and a BEV image for spatial layout. Since the two views are generated independently, they inevitably introduce cross-view conflicts. We therefore use semantic-physical knowledge priors, instantiated with a VLM, to reconcile these conflicts and obtain a unified procedure, \textbf{Prior-Guided Conflict-Aware Dual-View Initialization}.

From the generated front view and BEV, we detect object instances and 2D layout boxes, respectively. Because the two views are independently generated, they can disagree in object count, spatial placement, and appearance details. We resolve these conflicts by factorizing the cues. We therefore assign asymmetric roles to the two views. The front view is treated as the appearance and asset source, providing object identity, local visual details, and geometry references. The BEV is treated as the spatial source, providing ground-plane placement, scene topology, and coarse spatial support.

When the detected object counts are inconsistent between two views, we handle three cases. For matched objects, we instantiate the front-view asset and place it according to the corresponding BEV layout box. For BEV-only objects, we query the VLM with the prompt, neighboring objects, and current layout context. Objects that are inconsistent with the prompt or physically unsupported by the scene are removed or down-weighted as hallucinations, while plausible contextual elements are retained.
For front-only objects, we keep them as semantic anchors and use the VLM to infer tentative layout hypotheses from the prompt, object category, and nearby matched objects. These hypotheses are later refined together with other boxes in the joint optimization.

To lift BEV layouts into 3D placements, the VLM predicts a grounding prior $\gamma_i\in\{\mathrm{grounded},\mathrm{floating}\}$ for each retained object. Given a BEV box $\hat{r}_i^{\mathrm{BEV}}$ and an asset or category size prior $\bar{s}_i$, we initialize a 3D box by a grounding-aware lifting operator:
\begin{equation}
b_i^{(0)}=\operatorname{Lift}_{\gamma_i}(\hat{r}_i^{\mathrm{BEV}},\bar{s}_i).
\end{equation}
For grounded objects, the vertical center is placed according to the object height; for floating objects, the height is inferred from the VLM or category prior. This converts semantic grounding cues into physically plausible 3D placement without training a 3D detector.

Finally, we refine the retained boxes with a lightweight inference-time objective:
\begin{equation}
\mathcal{B}^{\star}=\arg\min_{\mathcal{B}}\mathcal{L}_{\mathrm{front}}+\lambda_b\mathcal{L}_{\mathrm{bev}}+\lambda_v\mathcal{L}_{\mathrm{vlm}}+\lambda_p\mathcal{L}_{\mathrm{phys}}.
\end{equation}
Here, $\mathcal{L}_{\mathrm{front}}$ preserves object identity and appearance cues from the front view, $\mathcal{L}_{\mathrm{bev}}$ preserves global layout cues from the BEV, $\mathcal{L}_{\mathrm{vlm}}$ enforces semantic existence and grounding priors, and $\mathcal{L}_{\mathrm{phys}}$ penalizes obvious physical violations such as collisions or invalid scene boundaries. The optimization updates only box parameters rather than model weights. This keeps the construction lightweight, training-free, and model-agnostic, while avoiding domain-specific 3D box detectors that typically require costly 3D annotations and can suffer from out-of-distribution failures when applied to open-ended generated scenes.

After box refinement, each optimized 3D box is used as the spatial pose term $p_i^0$. In parallel, the corresponding object is cropped from the front-view image and processed independently by an image-to-3D model to obtain its geometry $g_i^0$. Together with semantic attributes $s_i^0$ inferred from the input and knowledge priors, these terms form the initial world state:
\begin{equation}
\mathcal{W}^0=\{o_i^0\}_{i=1}^{N_0}, \qquad o_i^0=(g_i^0,p_i^0,s_i^0).
\end{equation}
The resulting $\mathcal{W}^0$ is an editable, spatially structured, and physically plausible initialization for subsequent state evolution and world interaction.

\begin{figure*}[th]
    \centering
    \includegraphics[width=1.0\linewidth]{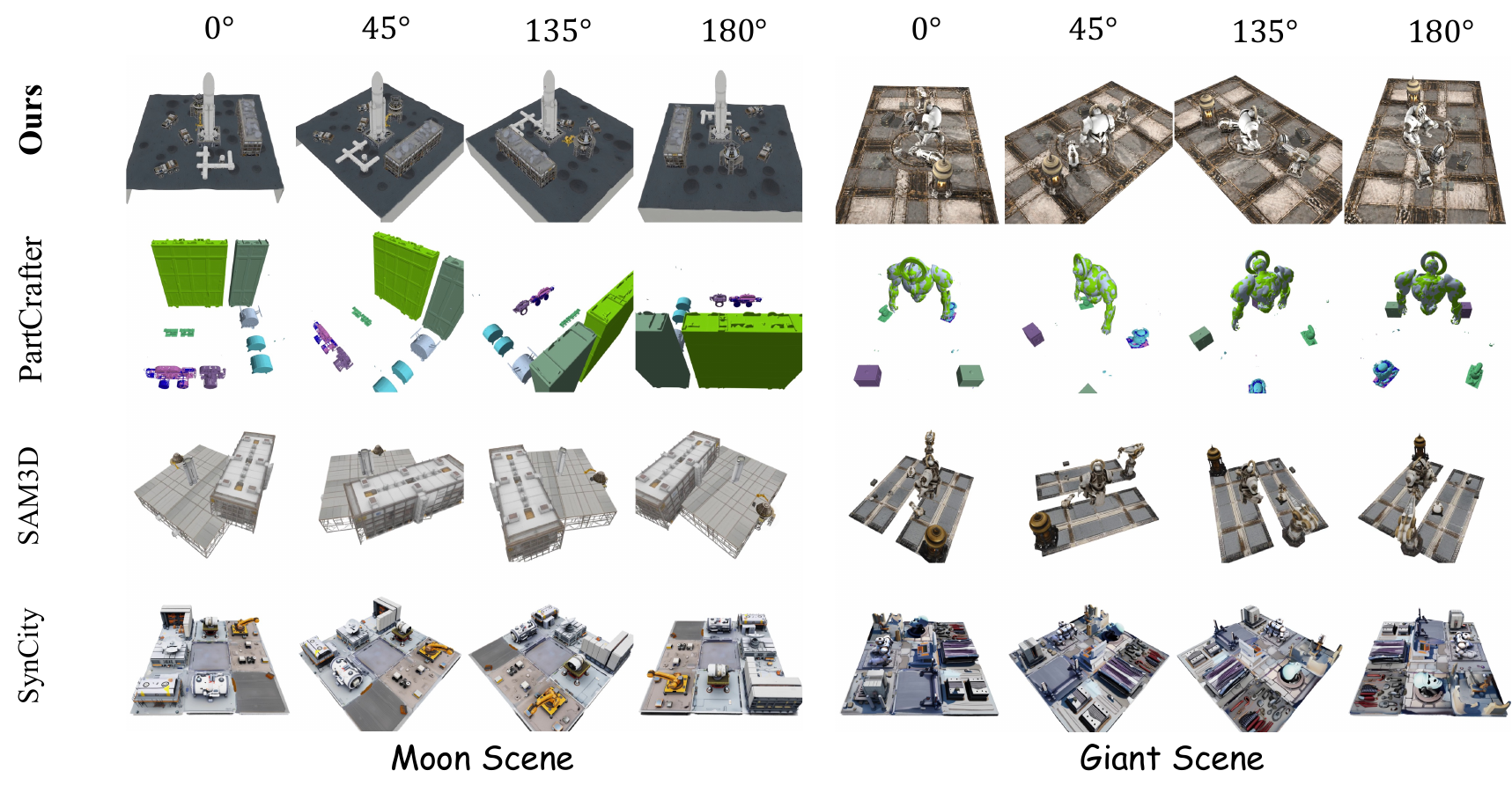}
    \caption{Qualitative comparison on Scene Generation}
    \label{fig:scene}
\end{figure*}

\subsection{State Evolution}

Given the initialized world state $\mathcal{W}^t$, the state evolution stage updates it into $\mathcal{W}^{t+1}$ according to user intent. Instead of asking a video renderer to implicitly hallucinate every dynamic process, we introduce \textbf{Intent-Guided Structured State Transition}, where a VLM queries the structured state table and predicts a compact transition plan specifying the objects or scene attributes to update, the transition type, and the target state.

At the scene level, the transition plan supports global evolution such as scene expansion and scene style change. For scene expansion, the system reuses the state construction procedure to instantiate new regions or objects, and then merges them into the existing world state while preserving the original spatial support and object identities. Scene style change is handled by rewriting over scene or object descriptors, allowing localized and consistent control of style and material attributes without regenerating the entire world.

At the object level, the plan performs category-aware updates over the object state $o_i^t=(g_i^t,p_i^t,s_i^t)$. Character evolution is mainly represented by pose or trajectory updates that preserve identity while expressing behavioral states. Rigid object motion update the position and pose term $p_i^t$ to represent movement. For events that are difficult to simulate continuously, such as explosion, destruction, or large non-rigid deformation, we use event-level asset substitution: the geometry term $g_i^t$ is replaced by a generated or retrieved target-state asset while preserving the object's identity, semantic role, and coarse placement. Appearance or status changes are handled by updating the semantic state term $s_i^t$.

Formally, the VLM produces a transition plan $\Delta_t$ from the current state and intent, and the next state is obtained by applying it to the structured state table:
\begin{equation}
\begin{aligned}
\Delta_t &= \operatorname{Plan}_{\mathrm{VLM}}(\mathcal{W}^t,u^t), \\
\mathcal{W}^{t+1}
&= \operatorname{Apply}(\mathcal{W}^t,\Delta_t)
 = \{o_i^{t+1}\}_{i=1}^{N_{t+1}}.
\end{aligned}
\end{equation}
Each updated object remains represented as $o_i^{t+1}=(g_i^{t+1},p_i^{t+1},s_i^{t+1})$. Depending on the transition type, the update affect the geometry term $g_i^t$, the position and pose term $p_i^t$, and the semantic state term $s_i^t$ either jointly or selectively. Because evolution selectively updates over persistent object records, the resulting $\mathcal{W}^{t+1}$ remains a structured and editable 3D world state, which can be further evolved over time or accessed by downstream camera trajectories.

\begin{table*}[h]
\centering
\caption{\textbf{Quantitative comparison with video generation methods on VBench.}
Higher values indicate better performance for all metrics.}
\label{tab:vbench_moon_results}
\setlength{\tabcolsep}{4pt}
\renewcommand{\arraystretch}{1.00}
\resizebox{\textwidth}{!}{%
\begin{tabular}{l|ccccccc}
\toprule
\textbf{Method} 
& \textbf{Subject Cons $\uparrow$} 
& \textbf{Background Cons $\uparrow$} 
& \textbf{Aesthetic $\uparrow$} 
& \textbf{Imaging $\uparrow$} 
& \textbf{Motion Smooth $\uparrow$} 
& \textbf{Flicker $\uparrow$} 
& \textbf{Average $\uparrow$} \\
\midrule

Animaker 
& 0.7509 
& 0.8343 
& \textbf{0.6795} 
& 0.6684 
& 0.9582 
& 0.9641 
& 0.8092 \\

MovieAgent 
& 0.7533 
& 0.8557 
& 0.5348 
& \textbf{0.7260} 
& 0.9895 
& 0.9858 
& 0.8075 \\

Wan2.2
& 0.8836 
& 0.9422 
& 0.5220 
& 0.6911 
& 0.9852  
& 0.9731 
& 0.8283 \\

Seedance2.0
& 0.8110
& 0.9329
& 0.6311
& 0.7184
& 0.9828
& 0.9565
& 0.8387 \\

\textbf{Ours} 
& \textbf{0.9135} 
& \textbf{0.9506} 
& 0.5257 
& 0.7185 
& \textbf{0.9923} 
& \textbf{0.9902} 
& \textbf{0.8484} \\

\bottomrule
\end{tabular}%
}
\end{table*}

\subsection{State Access}
Given the dynamic world state $\mathcal{W}^t$, the state access stage determines how the constructed world is used for downstream tasks. While a world state can be accessed through editing commands, physical simulation, or user control, camera trajectories provide a general interface for previsualization: they convert the same persistent world into film-like keyframes, explorable videos, and game-like viewpoints without modifying the world itself. However, previous text-only camera generation\cite{liu2024chatcam}  mainly relies on linguistic intent and lacks geometric feedback from the actual scene. To address this, we combine the semantic reasoning of a VLM with geometric evidence from rendered observations, leading to \textbf{World-State Camera Planning with Render-Feedback Reflection}.

We first read camera-relevant information from $\mathcal{W}^t$, including object identities, object poses, bounding boxes, collection hierarchy, important objects, and coarse scene extent. Canonical rendered views provide additional visual context. The user instruction is summarized into segment-level director intents $d_i$, which specify the desired target, shot style, and motion at a high level.

For each segment, the VLM acts as a semantic proposer rather than the final controller. It generates an initial trajectory from the world state, rendered scene observations $V$, and director intent:
\begin{equation}
\pi_i^0=\operatorname{Propose}_{\mathrm{VLM}}(\mathcal{W}^t,V,d_i).
\end{equation}
We then execute the proposal in the 3D world through low-cost rendering and evaluate the rendered result:
\begin{equation}
R_i^k=\operatorname{Render}(\mathcal{W}^t,\pi_i^k), \qquad
 e_i^k=\operatorname{Eval}(R_i^k,\pi_i^k,d_i,\mathcal{W}^t).
\end{equation}
Here, $e_i^k$ summarizes discrepancies such as intent mismatch, target invisibility, poor framing, collision risk, and motion artifacts.

Reflection converts these discrepancies into local repair candidates over camera parameters. For example, visibility or framing issues can be repaired by adjusting camera distance, look-at targets, or field of view; occlusion and collision issues can be repaired by changing viewpoint height, azimuth, or moving along collision-free directions. Let $\Delta_m$ denote the $m$-th local repair operation applied to the current trajectory. These repairs generate a compact candidate set around $\pi_i^k$ rather than performing global search, from which we select the best trajectory:
\begin{equation}
\mathcal{P}_i^k=\{\pi_i^k+\Delta_m\}_{m=1}^{M}, \qquad
\pi_i^{k+1}=\arg\min_{\pi\in\mathcal{P}_i^k}J(\pi;d_i,\mathcal{W}^t,R_i^k).
\end{equation}
The score $J$ favors better trajectories satisfy intent alignment, target visibility, framing quality, collision avoidance, and motion smoothness. This local proposal-and-verification loop is efficient because it searches only over issue-driven repairs, and iterates until no significant issue is detected or maximum rounds is reached. Since only camera parameters are updated, the planner remains training-free and does not require camera trajectory annotations.
The final trajectory $\pi_i^\star$ provides state access for keyframe generation, free-view exploration, and third-person views while remaining grounded in the same persistent 3D world state.

\section{Experiments}

\subsection{Implementation Details.}
Our system takes a single text prompt from the user as input and automatically completes the full workflow from 3D world construction to downstream application export. We use Gemini 3.1~\cite{google2026gemini31pro} as the vision-language model for scene understanding, intent parsing, layout reasoning, state transition planning, and camera feedback reflection. For image generation, we use Nano Banana 2~\cite{google2026nanobanana2} to synthesize the front-view and bird's-eye-view references used in state construction. Object-centric image crops are lifted into 3D assets using the latest Hunyuan3D model~\cite{lai2025hunyuan3d25highfidelity3d}. For video-based downstream outputs, we use Seedance2~\cite{seedance2026seedance} as the video generation backbone.

\begin{table}[h]
\centering
\caption{\textbf{Quantitative comparison with scene generation.}
We report CLIP-I, CLIP-T, HPS V2 and Q-Align. 
Higher values indicate better performance.}
\label{tab:results}
\setlength{\tabcolsep}{10pt}
\renewcommand{\arraystretch}{0.9}
\resizebox{0.5\textwidth}{!}{%
\begin{tabular}{l|ccc|c}
\toprule
\textbf{Method} 
& \textbf{CLIP-I $\uparrow$} 
& \textbf{CLIP-T $\uparrow$} 
& \textbf{HPS V2 $\uparrow$} 
& \textbf{Q-Align Quality/Aesthetics $\uparrow$} 

\\
\midrule
PartCrafter 
& 0.542
& 20.761 
& 0.110 
& 1.675/1.587 
 \\
SAM3D 
& 0.580 
& 15.481 
& 0.055 
& 2.262/2.137 
 \\
SynCity 
& 0.689 
& 22.880 
& \textbf{0.175} 
& 3.535/\textbf{2.966} 
 \\
\textbf{Ours} 
& \textbf{0.788} 
& \textbf{30.214} 
& 0.151
& \textbf{3.621}/2.859
\\
\bottomrule
\end{tabular}%
}
\end{table}

\begin{figure*}[p]
  \centering
\includegraphics[width=1.0\textwidth,page=1]{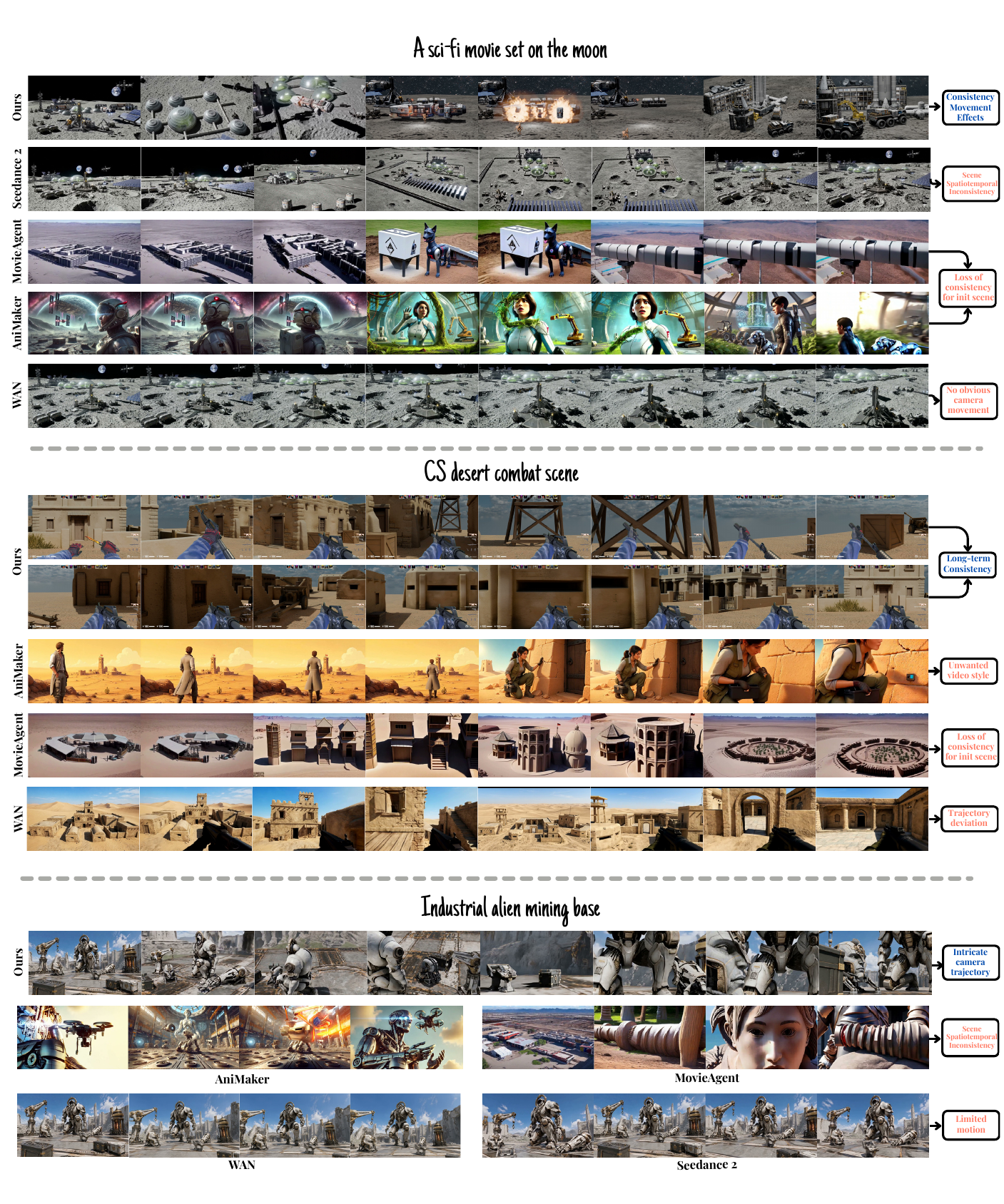}
  \caption{Qualitative comparison on previsualization applications.
  Case~1 and Case~3 showcase \textbf{video creation}, while Case~2 shows \textbf{3D game prototyping}.
  Video-generation baselines consistently suffer from spatial--temporal inconsistency, identity drifting, and restricted camera motion, whereas StateFlow operates on a persistent and interactive 3D world, yielding precise, highly controllable, and geometry--appearance-consistent results.}
  \label{fig:supp1}
\end{figure*}

\newcommand{\metrichead}[2]{\makecell[c]{\textbf{#1}\\\textbf{#2} $\uparrow$}}

\begin{table*}[th]
\centering
\caption{
User study and MLLM-based evaluation on 3D scene generation (top) and video generation (bottom).
Each cell is reported as \textit{user study / MLLM evaluation}.
User study: $N{=}30$ participants, $12$ prompts per setting.
\textbf{Bold} represents the best performance.
}
\label{tab:user_study}

\small
\setlength{\tabcolsep}{4.8pt}
\renewcommand{\arraystretch}{1.18}

\begin{adjustbox}{max width=\textwidth}
\begin{tabular}{l@{\hspace{0.9em}}ccccccc}
\toprule
\multicolumn{8}{l}{\emph{Scene-level study (}$45^\circ$\emph{ BEV fly-around videos)}} \\
\midrule
\textbf{Method}
& \metrichead{Prompt}{Align.}
& \metrichead{Layout}{Plaus.}
& \metrichead{Complete}{ness}
& \metrichead{Geo.}{Quality}
& \metrichead{Coher.}{ence}
& \metrichead{Previs}{Useful.}
& \metrichead{Overall}{Score} \\
\midrule
SynCity                   & 3.5/2.8 & 2.8/2.8 & 3.1/2.6 & 3.9/2.8 & 2.9/4.4 & 2.7/2.6 & 3.2/3.0 \\
SAM3D                     & 3.9/2.4 & 3.1/2.6 & 3.7/2.2 & 3.9/2.2 & 3.2/4.6 & 3.1/1.2 & 3.5/2.5 \\
PartCrafter               & 2.4/2.0 & 1.9/2.0 & 1.9/2.0 & 2.2/2.0 & 2.0/3.2 & 1.8/1.0 & 2.0/2.0 \\
w/o BEV Layout (Ours)     & 3.6/3.0 & 3.0/2.6 & 3.4/2.0 & 3.7/2.4 & 3.0/3.6 & 3.1/1.2 & 3.3/2.5 \\
w/o Conflict Res. (Ours)  & 3.7/3.0 & 3.2/2.6 & 3.2/2.0 & 3.8/2.6 & 3.1/3.6 & 3.2/1.6 & 3.4/2.6 \\
\textbf{StateFlow (Ours)} & \textbf{4.4}/\textbf{3.6} & \textbf{4.6}/\textbf{3.3} & \textbf{4.2}/\textbf{3.7} & \textbf{4.2}/\textbf{3.6} & \textbf{4.6}/\textbf{4.7} & \textbf{4.7}/\textbf{3.1} & \textbf{4.5}/\textbf{3.7} \\
\midrule
\multicolumn{8}{l}{\emph{Video-level study (final rendered videos)}} \\
\midrule
\textbf{Method}
& \metrichead{Prompt/Story}{Align.}
& \metrichead{Spatial}{Cons.}
& \metrichead{Identity}{Cons.}
& \metrichead{Motion}{Plaus.}
& \metrichead{Camera}{Quality}
& \metrichead{Previs}{Useful.}
& \metrichead{Overall}{Score} \\
\midrule
Seedance2                 & 4.1/3.4 & 3.3/1.5 & 3.2/1.3 & 3.9/2.6 & 3.5/1.4 & 3.7/2.3 & 3.6/2.1 \\
Wan2.2                    & 3.5/3.6 & 2.8/2.4 & 2.7/2.2 & 3.4/2.7 & 3.0/1.6 & 3.1/1.5 & 3.1/2.3 \\
AniMaker                  & 2.8/1.6 & 2.3/1.2 & 2.2/1.5 & 2.8/2.4 & 2.5/2.3 & 2.4/1.4 & 2.5/1.7 \\
MovieAgent                & 1.9/1.3 & 1.6/1.1 & 1.5/1.2 & 1.9/2.2 & 1.6/1.4 & 1.5/1.3 & 1.7/1.4 \\
VLM-only Camera (Ours)    & 4.3/3.5 & 4.6/3.4 & 4.6/3.7 & 4.1/3.2 & 3.4/2.6 & 3.8/3.5 & 4.1/3.3 \\
StateFlow (Ours)          & \textbf{4.4}/\textbf{3.8} & \textbf{4.7}/\textbf{4.3} & \textbf{4.7}/\textbf{3.9} & \textbf{4.2}/\textbf{3.6} & \textbf{4.6}/\textbf{4.4} & \textbf{4.6}/\textbf{4.2} & \textbf{4.5}/\textbf{4.0} \\
\bottomrule
\end{tabular}
\end{adjustbox}

\vspace{-2mm}
\end{table*}

\subsection{Comparison on Scene Generation}
We compare StateFlow with representative 3D generation baselines covering different paradigms. SynCity~\cite{engstler2025syncity} generates text-conditioned 3D worlds in a training-free, tile-by-tile manner, SAM3D~\cite{sam3dteam2025sam3d3dfyimages} represents feed-forward image-to-3D generation, and PartCrafter~\cite{lin2025partcrafter} represents part-aware 3D object generation. Since our wild inference examples do not have ground-truth 3D scenes, we render each generated result into a video from a $45^\circ$ bird's-eye-view camera and compute CLIP-I and CLIP-T~\cite{radford2021learning}, Human Preference Score (HPS)~\cite{wu2023human} and Q-Align score~\cite{wu2023q} to measure visual and text alignment. As shown in Table~\ref{tab:results}, the results show that our method achieves overall superior performance, with particularly clear advantages in CLIP-I and CLIP-T. We note that HPS and Q-Align tend to exhibit strong preferences toward colorfulness and visual style, which may lead to relatively high scores for SynCity.

We also provide qualitative comparisons in Fig.~\ref{fig:scene} to inspect whether the generated outputs provide coherent spatial layout and editable scene structure required by previsualization. The visual results show that StateFlow produces more coherent and fine-grained 3D worlds than the compared baselines. In particular, our method better preserves global spatial organization while maintaining object-level details, resulting in scenes that are more suitable for downstream previsualization. In contrast, the baseline methods often suffer from incomplete scene construction, weak object relationships, or limited editable structure.

\subsection{Comparison on Video Creation}
We compare StateFlow with video generation methods, such as Animaker~\cite{shi2025animaker}, MovieAgent~\cite{wu2025automated}, Wan2.2~\cite{wan2025wan} and Seedance2.0~\cite{seedance2026seedance}. These methods span large-scale text-to-video generation, narrative-driven animation synthesis, and agentic multi-shot video generation, providing a broad comparison against recent video paradigms.

Table~\ref{tab:vbench_moon_results} reports the quantitative comparison with representative video generation methods using VBench~\cite{huang2024vbench}. Our method achieves the best average score among all compared methods, indicating its overall advantage in video generation quality. In particular, our method obtains the highest scores in subject consistency, background consistency, motion smoothness, and flicker, suggesting that the generated videos better preserve the main object, maintain stable scene content, and exhibit smoother temporal dynamics. Although some baselines achieve slightly higher scores on aesthetic or imaging quality, these metrics are more sensitive to low-level visual preferences such as colorfulness and contrast. In contrast, our method shows stronger performance on structure and temporality related dimensions, which are more critical for controllable scene generation and previsualization.

Beyond these quantitative results, Fig.~\ref{fig:supp1} presents a visual comparison with these video generation methods. StateFlow produces outputs with more stable scene structure and clearer spatial grounding, especially when the camera moves or the scene is observed from changing viewpoints. The generated videos preserve the major objects and environmental layout more reliably, while avoiding severe temporal drift or background inconsistency.

\begin{figure*}[th]
    \centering
    \includegraphics[width=1.0\linewidth]{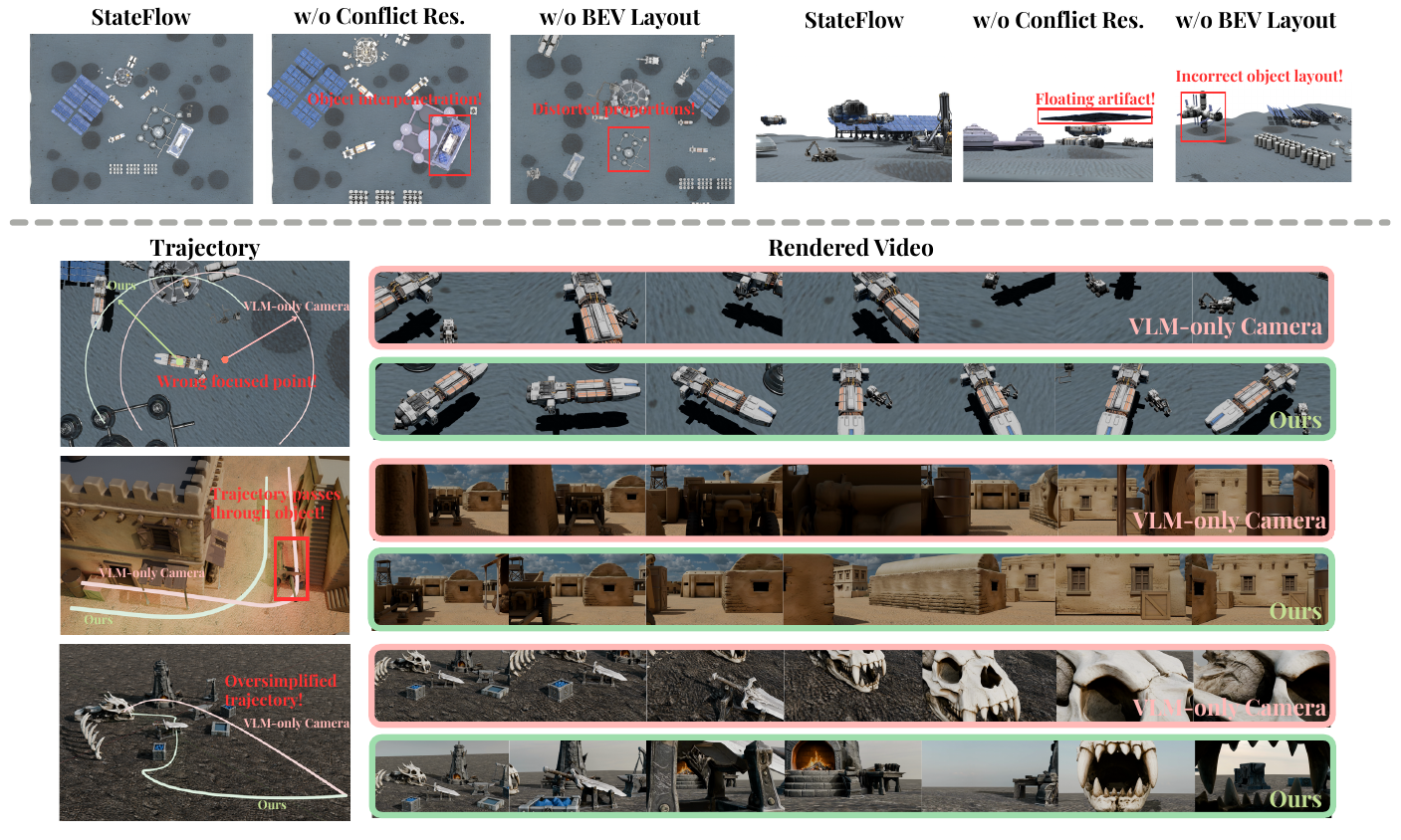}
    \caption{Visualization of Ablation Study}
    \label{fig:abl_vis}
\end{figure*}

\subsection{MLLM-based Evaluation and User Study}
We further conduct both MLLM-based evaluation and user study to assess perceptual quality and practical usefulness, as summarized in Table~\ref{tab:user_study}. The evaluation covers two settings: a \emph{scene-level} study, where each method's output is rendered as a $45^\circ$ bird's-eye-view fly-around video, and a \emph{video-level} study, where the final rendered videos are evaluated. For each prompt, results from all methods are shown side by side in anonymized and randomized order. Participants rate each method on a 5-point Likert scale, where 1 indicates very poor quality and 5 indicates excellent quality. We also clarify that a score of 5 does not require perfection, allowing multiple methods to receive high scores when appropriate. We recruit $N{=}30$ participants with $12$ prompts for each setting, insert attention-check items, and report per-dimension mean scores. In addition, we use gemini-3.1 as an MLLM evaluation to provide complementary automatic assessment.

At the scene level, StateFlow achieves the best performance across all evaluated dimensions, including prompt alignment, layout plausibility, completeness, geometric quality, coherence, previsualization usefulness, and overall quality. These results indicate that the proposed world-state construction produces more complete, spatially plausible, and coherent 3D scenes than existing baselines. 

At the video level, StateFlow again achieves the highest overall score in both the user study and MLLM-based evaluation. Compared with video generation baselines, our method shows clear advantages in spatial consistency, identity consistency, camera quality, and previsualization usefulness. This suggests that explicitly constructing and accessing a persistent 3D world better preserves object identity and scene layout across time and viewpoints, while supporting more controllable camera trajectories. Although some video baselines achieve competitive prompt/story alignment, they often exhibit weaker spatial consistency and less reliable camera behavior. In contrast, StateFlow produces more coherent and production-oriented results, making it more suitable for previsualization workflows.

\subsection{Ablation Study}
We further use the same MLLM-based evaluation and user study protocol to validate the key design choices of StateFlow, since these components mainly affect structural coherence, editability, and camera controllability, which are difficult to capture with metrics. The results are reported alongside the baselines in Tab.~\ref{tab:user_study}.
\paragraph{State Construction.}
We evaluate two construction ablations in the scene-level study. \emph{w/o BEV Layout} removes the top-down spatial grounding and relies mainly on appearance-centric cues, which tends to produce weaker global layout, incorrect object scale, and less coherent object relations. \emph{w/o Conflict Res.} directly combines front-view and BEV cues without resolving object-count and spatial mismatches, leading to missing, duplicated, or implausibly placed objects. These variants verify the importance of BEV grounding and conflict-aware reconciliation for building a coherent 3D world.
\paragraph{State Evolution.}
We additionally ablate the structured state transition by replacing selective state updates with direct regeneration for edited content. This weakens identity preservation and spatial consistency across edits, since the system no longer updates persistent object records. This validates the need for evolving the world through explicit state changes rather than treating each edit as an independent generation task.
\paragraph{State Access.}
For camera planning, we compare against \emph{VLM-only Camera}, which generates trajectories from semantic intent without render-feedback reflection. This often causes visibility, occlusion, framing, or collision issues because semantic reasoning alone cannot fully predict the visual outcome in the concrete 3D scene. The comparison verifies the effectiveness of using rendered feedback to refine camera trajectories.

\section{Conclusion}
This paper introduces \textbf{StateFlow}, a state-centric framework for previsualization. Rather than generating videos in one shot, it uses an editable 3D world to organize scene structure, evolution, and cameras, while off-the-shelf video models can enhance visual quality without altering the underlying world. We formulate previsualization around this persistent structured 3D state, composed of object geometry, spatial pose, and semantic attributes. Based on this representation, StateFlow initializes 3D worlds through prior-guided dual-view construction, evolves them via intent-guided state transitions, and accesses them through render-feedback camera planning for downstream applications such as keyframe generation, explorable observation, video creation, and third-person interaction.

Extensive experiments demonstrate that StateFlow provides an effective and versatile foundation for controllable previsualization, producing visually faithful results while maintaining coherent spatial layout, stable scene structure, and editable object-level organization. Beyond improving generation quality, our framework highlights the value of explicit 3D world states as an intermediate representation between generative models and production-oriented creative workflows. We believe StateFlow opens a promising direction toward more interactive, structured, and world-aware generative systems for future content creation.

\paragraph{Limitation and Future Work}
Although our method accelerates previsualization from the weeks or months typically required in industrial workflows to several minutes, it is still limited by the inference speed of third-party models and therefore cannot yet support fully real-time interaction. In the future, with more efficient deployment and faster inference of the associated models, our method can be further accelerated.

{
    \small
    \bibliographystyle{ieeenat_fullname}
    \bibliography{ref}

@inproceedings{nitsche2008experiments,
  title={Experiments in the use of game technology for pre-visualization},
  author={Nitsche, Michael},
  booktitle={Proceedings of the 2008 Conference on Future Play: Research, Play, Share},
  pages={160--165},
  year={2008}
}

@incollection{northam2012collaborative,
  title={A collaborative real time previsualization tool for video games and film},
  author={Northam, Lesley and Istead, Joe and Kaplan, Craig S},
  booktitle={ACM SIGGRAPH 2012 Posters},
  pages={1--1},
  year={2012}
}

@inproceedings{ardal2019collaborative,
  title={A collaborative previsualization tool for filmmaking in virtual reality},
  author={Ardal, Dui and Alexandersson, Simon and Lempert, Mirko and Abelho Pereira, Andr{\'e} Tiago},
  booktitle={Proceedings of the 16th ACM SIGGRAPH European Conference on Visual Media Production},
  pages={1--10},
  year={2019}
}

@inproceedings{fei2023tape,
  title={TAPE: Tangible Augmented Previz Environment for Filmmaking},
  author={Fei, Guangzheng and Liu, Dake},
  booktitle={Proceedings of the Eleventh International Symposium of Chinese CHI},
  pages={251--262},
  year={2023}
}

@inproceedings{jo2024collagevis,
  title={CollageVis: Rapid Previsualization Tool for Indie Filmmaking using Video Collages},
  author={Jo, Hye-Young and Suzuki, Ryo and Kim, Yoonji},
  booktitle={Proceedings of the 2024 CHI Conference on Human Factors in Computing Systems},
  pages={1--16},
  year={2024}
}

@inproceedings{ivanov2022one,
  title={One week in the future: Previs design futuring for HCI research},
  author={Ivanov, Alexander and Au Yeung, Tim and Blair, Kathryn and Danyluk, Kurtis and Freeman, Georgina and Friedel, Marcus and Hull, Carmen and Hung, Michael Yuk-Shing and Pratte, Sydney and Willett, Wesley},
  booktitle={Proceedings of the 2022 CHI Conference on Human Factors in Computing Systems},
  pages={1--15},
  year={2022}
}

@article{yu2022systematic,
  title={A systematic review of architectural design collaboration in immersive virtual environments},
  author={Yu, Rongrong and Gu, Ning and Lee, Gun and Khan, Ayaz},
  journal={Designs},
  volume={6},
  number={5},
  pages={93},
  year={2022},
  publisher={MDPI}
}

@inproceedings{christie2008camera,
  title={Camera control in computer graphics},
  author={Christie, Marc and Olivier, Patrick and Normand, Jean-Marie},
  booktitle={Computer graphics forum},
  volume={27},
  number={8},
  pages={2197--2218},
  year={2008},
  organization={Wiley Online Library}
}

@inproceedings{halperin2025camera,
  title={From camera-eye to AI: Exploring the interplay of cinematography and computational visual storytelling},
  author={Halperin, Brett A and Lukin, Stephanie M},
  booktitle={Proceedings of the 2025 CHI Conference on Human Factors in Computing Systems},
  pages={1--18},
  year={2025}
}

@inproceedings{lino2011director,
  title={The director's lens: an intelligent assistant for virtual cinematography},
  author={Lino, Christophe and Christie, Marc and Ranon, Roberto and Bares, William},
  booktitle={Proceedings of the 19th ACM international conference on Multimedia},
  pages={323--332},
  year={2011}
}

@inproceedings{rombach2022high,
  title={High-resolution image synthesis with latent diffusion models},
  author={Rombach, Robin and Blattmann, Andreas and Lorenz, Dominik and Esser, Patrick and Ommer, Bj{\"o}rn},
  booktitle={Proceedings of the IEEE/CVF conference on computer vision and pattern recognition},
  pages={10684--10695},
  year={2022}
}

@article{brooks2024video,
  title={Video generation models as world simulators},
  author={Brooks, Tim and Peebles, Bill and Holmes, Connor and DePue, Will and Guo, Yufei and Jing, Leo and Schnurr, David and Taylor, Joe and Luhman, Troy and Luhman, Eric and others},
  journal={OpenAI Blog},
  volume={1},
  number={8},
  pages={1},
  year={2024}
}

@article{yang2024cogvideox,
  title={Cogvideox: Text-to-video diffusion models with an expert transformer},
  author={Yang, Zhuoyi and Teng, Jiayan and Zheng, Wendi and Ding, Ming and Huang, Shiyu and Xu, Jiazheng and Yang, Yuanming and Hong, Wenyi and Zhang, Xiaohan and Feng, Guanyu and others},
  journal={arXiv preprint arXiv:2408.06072},
  year={2024}
}

@article{blattmann2023stable,
  title={Stable video diffusion: Scaling latent video diffusion models to large datasets},
  author={Blattmann, Andreas and Dockhorn, Tim and Kulal, Sumith and Mendelevitch, Daniel and Kilian, Maciej and Lorenz, Dominik and Levi, Yam and English, Zion and Voleti, Vikram and Letts, Adam and others},
  journal={arXiv preprint arXiv:2311.15127},
  year={2023}
}

@article{kong2024hunyuanvideo,
  title={Hunyuanvideo: A systematic framework for large video generative models},
  author={Kong, Weijie and Tian, Qi and Zhang, Zijian and Min, Rox and Dai, Zuozhuo and Zhou, Jin and Xiong, Jiangfeng and Li, Xin and Wu, Bo and Zhang, Jianwei and others},
  journal={arXiv preprint arXiv:2412.03603},
  year={2024}
}

@article{wan2025wan,
  title={Wan: Open and advanced large-scale video generative models},
  author={Wan, Team and Wang, Ang and Ai, Baole and Wen, Bin and Mao, Chaojie and Xie, Chen-Wei and Chen, Di and Yu, Feiwu and Zhao, Haiming and Yang, Jianxiao and others},
  journal={arXiv preprint arXiv:2503.20314},
  year={2025}
}

@inproceedings{bruce2024genie,
  title={Genie: Generative interactive environments},
  author={Bruce, Jake and Dennis, Michael D and Edwards, Ashley and Parker-Holder, Jack and Shi, Yuge and Hughes, Edward and Lai, Matthew and Mavalankar, Aditi and Steigerwald, Richie and Apps, Chris and others},
  booktitle={Forty-first International Conference on Machine Learning},
  year={2024}
}

@inproceedings{xiang2025structured,
  title={Structured 3d latents for scalable and versatile 3d generation},
  author={Xiang, Jianfeng and Lv, Zelong and Xu, Sicheng and Deng, Yu and Wang, Ruicheng and Zhang, Bowen and Chen, Dong and Tong, Xin and Yang, Jiaolong},
  booktitle={2025 IEEE/CVF Conference on Computer Vision and Pattern Recognition (CVPR)},
  pages={21469--21480},
  year={2025},
  organization={IEEE}
}

@article{zhao2025hunyuan3d,
  title={Hunyuan3d 2.0: Scaling diffusion models for high resolution textured 3d assets generation},
  author={Zhao, Zibo and Lai, Zeqiang and Lin, Qingxiang and Zhao, Yunfei and Liu, Haolin and Yang, Shuhui and Feng, Yifei and Yang, Mingxin and Zhang, Sheng and Yang, Xianghui and others},
  journal={arXiv preprint arXiv:2501.12202},
  year={2025}
}

@misc{lai2025hunyuan3d25highfidelity3d,
      title={Hunyuan3D 2.5: Towards High-Fidelity 3D Assets Generation with Ultimate Details}, 
      author={Tencent Hunyuan3D Team},
      year={2025},
      eprint={2506.16504},
      archivePrefix={arXiv},
      primaryClass={cs.CV},
      url={https://arxiv.org/abs/2506.16504}, 
}

@article{sun2024dimensionx,
  title={Dimensionx: Create any 3d and 4d scenes from a single image with controllable video diffusion},
  author={Sun, Wenqiang and Chen, Shuo and Liu, Fangfu and Chen, Zilong and Duan, Yueqi and Zhang, Jun and Wang, Yikai},
  journal={arXiv preprint arXiv:2411.04928},
  year={2024}
}

@article{sun2025worldplay,
  title={Worldplay: Towards long-term geometric consistency for real-time interactive world modeling},
  author={Sun, Wenqiang and Zhang, Haiyu and Wang, Haoyuan and Wu, Junta and Wang, Zehan and Wang, Zhenwei and Wang, Yunhong and Zhang, Jun and Wang, Tengfei and Guo, Chunchao},
  journal={arXiv preprint arXiv:2512.14614},
  year={2025}
}

@article{zhang2026worldstereo,
  title={WorldStereo: Bridging Camera-Guided Video Generation and Scene Reconstruction via 3D Geometric Memories},
  author={Zhang, Yisu and Cao, Chenjie and Wang, Tengfei and Zuo, Xuhui and Wu, Junta and Zhu, Jianke and Guo, Chunchao},
  journal={arXiv preprint arXiv:2603.02049},
  year={2026}
}

@article{liu2024chatcam, title={Chatcam: Empowering camera control through conversational ai}, author={Liu, Xinhang and Tai, Yu-Wing and Tang, Chi-Keung}, journal={Advances in Neural Information Processing Systems}, volume={37}, pages={54483--54506}, year={2024} }

@inproceedings{infinigen2023infinite,
  title={Infinite Photorealistic Worlds Using Procedural Generation},
  author={Raistrick, Alexander and Lipson, Lahav and Ma, Zeyu and Mei, Lingjie and Wang, Mingzhe and Zuo, Yiming and Kayan, Karhan and Wen, Hongyu and Han, Beining and Wang, Yihan and Newell, Alejandro and Law, Hei and Goyal, Ankit and Yang, Kaiyu and Deng, Jia},
  booktitle={Proceedings of the IEEE/CVF Conference on Computer Vision and Pattern Recognition},
  pages={12630--12641},
  year={2023}
}

@inproceedings{infinigen2024indoors,
    author    = {Raistrick, Alexander and Mei, Lingjie and Kayan, Karhan and Yan, David and Zuo, Yiming and Han, Beining and Wen, Hongyu and Parakh, Meenal and Alexandropoulos, Stamatis and Lipson, Lahav and Ma, Zeyu and Deng, Jia},
    title     = {Infinigen Indoors: Photorealistic Indoor Scenes using Procedural Generation},
    booktitle = {Proceedings of the IEEE/CVF Conference on Computer Vision and Pattern Recognition (CVPR)},
    month     = {June},
    year      = {2024},
    pages     = {21783-21794}
}

@inproceedings{yang2024holodeck,
  title={Holodeck: Language guided generation of 3d embodied ai environments},
  author={Yang, Yue and Sun, Fan-Yun and Weihs, Luca and VanderBilt, Eli and Herrasti, Alvaro and Han, Winson and Wu, Jiajun and Haber, Nick and Krishna, Ranjay and Liu, Lingjie and others},
  booktitle={Proceedings of the IEEE/CVF Conference on Computer Vision and Pattern Recognition},
  pages={16227--16237},
  year={2024}
}

@article{yang2024llplace,
  title={Llplace: The 3d indoor scene layout generation and editing via large language model},
  author={Yang, Yixuan and Lu, Junru and Zhao, Zixiang and Luo, Zhen and Yu, James JQ and Sanchez, Victor and Zheng, Feng},
  journal={arXiv preprint arXiv:2406.03866},
  year={2024}
}

@inproceedings{sun2025layoutvlm,
  title={Layoutvlm: Differentiable optimization of 3d layout via vision-language models},
  author={Sun, Fan-Yun and Liu, Weiyu and Gu, Siyi and Lim, Dylan and Bhat, Goutam and Tombari, Federico and Li, Manling and Haber, Nick and Wu, Jiajun},
  booktitle={Proceedings of the Computer Vision and Pattern Recognition Conference},
  pages={29469--29478},
  year={2025}
}

@article{feng2023layoutgpt,
  title={Layoutgpt: Compositional visual planning and generation with large language models},
  author={Feng, Weixi and Zhu, Wanrong and Fu, Tsu-jui and Jampani, Varun and Akula, Arjun and He, Xuehai and Basu, Sugato and Wang, Xin Eric and Wang, William Yang},
  journal={Advances in Neural Information Processing Systems},
  volume={36},
  pages={18225--18250},
  year={2023}
}

@inproceedings{gu2025artiscene,
  title={Artiscene: Language-driven artistic 3d scene generation through image intermediary},
  author={Gu, Zeqi and Cui, Yin and Li, Zhaoshuo and Wei, Fangyin and Ge, Yunhao and Gu, Jinwei and Liu, Ming-Yu and Davis, Abe and Ding, Yifan},
  booktitle={Proceedings of the Computer Vision and Pattern Recognition Conference},
  pages={2891--2901},
  year={2025}
}

@article{zhu2025imaginarium,
  title={Imaginarium: Vision-guided High-Quality 3D Scene Layout Generation},
  author={Zhu, Xiaoming and Huang, Xu and Xie, Qinghongbing and Deng, Zhi and Yu, Junsheng and Guan, Yirui and Liu, Zhongyuan and Zhu, Lin and Zhao, Qijun and Liu, Ligang and others},
  journal={ACM Transactions on Graphics (TOG)},
  volume={44},
  number={6},
  pages={1--24},
  year={2025},
  publisher={ACM New York, NY, USA}
}

@article{sam3dteam2025sam3d3dfyimages,
      title={SAM 3D: 3Dfy Anything in Images}, 
      author={SAM 3D Team and Xingyu Chen and Fu-Jen Chu and Pierre Gleize and Kevin J Liang and Alexander Sax and Hao Tang and Weiyao Wang and Michelle Guo and Thibaut Hardin and Xiang Li and Aohan Lin and Jiawei Liu and Ziqi Ma and Anushka Sagar and Bowen Song and Xiaodong Wang and Jianing Yang and Bowen Zhang and Piotr Dollár and Georgia Gkioxari and Matt Feiszli and Jitendra Malik},
      year={2025},
      eprint={2511.16624},
      archivePrefix={arXiv},
      primaryClass={cs.CV},
      url={https://arxiv.org/abs/2511.16624}, 
}

@article{wang2025worldgen,
  title={WorldGen: From Text to Traversable and Interactive 3D Worlds},
  author={Wang, Dilin and Jung, Hyunyoung and Monnier, Tom and Sohn, Kihyuk and Zou, Chuhang and Xiang, Xiaoyu and Yeh, Yu-Ying and Liu, Di and Huang, Zixuan and Nguyen-Phuoc, Thu and others},
  journal={arXiv preprint arXiv:2511.16825},
  year={2025}
}

@article{shi2025scenemaker,
  title={SceneMaker: Open-set 3D Scene Generation with Decoupled De-occlusion and Pose Estimation Model},
  author={Shi, Yukai and Li, Weiyu and Wang, Zihao and Li, Hongyang and Chen, Xingyu and Tan, Ping and Zhang, Lei},
  journal={arXiv preprint arXiv:2512.10957},
  year={2025}
}

@article{lin2025partcrafter,
  title={Partcrafter: Structured 3d mesh generation via compositional latent diffusion transformers},
  author={Lin, Yuchen and Lin, Chenguo and Pan, Panwang and Yan, Honglei and Feng, Yiqiang and Mu, Yadong and Fragkiadaki, Katerina},
  journal={arXiv preprint arXiv:2506.05573},
  year={2025}
}

@inproceedings{huang2025midi,
  title={Midi: Multi-instance diffusion for single image to 3d scene generation},
  author={Huang, Zehuan and Guo, Yuan-Chen and An, Xingqiao and Yang, Yunhan and Li, Yangguang and Zou, Zi-Xin and Liang, Ding and Liu, Xihui and Cao, Yan-Pei and Sheng, Lu},
  booktitle={Proceedings of the IEEE/CVF Conference on Computer Vision and Pattern Recognition},
  pages={23646--23657},
  year={2025}
}

@article{yang2026flowscene,
  title={FlowScene: Style-Consistent Indoor Scene Generation with Multimodal Graph Rectified Flow},
  author={Yang, Zhifei and Zhai, Guangyao and Lu, Keyang and Yin, YuYang and Zhang, Chao and Xiao, Zhen and Long, Jieyi and Navab, Nassir and Wang, Yikai},
  journal={arXiv preprint arXiv:2603.19598},
  year={2026}
}

@inproceedings{yang2025mmgdreamer,
  title={Mmgdreamer: Mixed-modality graph for geometry-controllable 3d indoor scene generation},
  author={Yang, Zhifei and Lu, Keyang and Zhang, Chao and Qi, Jiaxing and Jiang, Hanqi and Ma, Ruifei and Yin, Shenglin and Xu, Yifan and Xing, Mingzhe and Xiao, Zhen and others},
  booktitle={Proceedings of the AAAI Conference on Artificial Intelligence},
  volume={39},
  number={9},
  pages={9391--9399},
  year={2025}
}

@inproceedings{engstler2025syncity,
  title={Syncity: Training-free generation of 3d worlds},
  author={Engstler, Paul and Shtedritski, Aleksandar and Laina, Iro and Rupprecht, Christian and Vedaldi, Andrea},
  booktitle={Proceedings of the IEEE/CVF International Conference on Computer Vision},
  pages={27585--27595},
  year={2025}
}

@misc{google2026gemini31pro,
  author       = {{Google}},
  title        = {{Gemini 3.1 Pro Preview}},
  year         = {2026},
  howpublished = {\url{https://ai.google.dev/gemini-api/docs/models/gemini-3.1-pro-preview}},
  note         = {Accessed: 2026-05-12}
}

@misc{google2026nanobanana2,
  author       = {{Google}},
  title        = {{Nano Banana 2: Google's latest AI image generation model}},
  year         = {2026},
  howpublished = {\url{https://blog.google/innovation-and-ai/technology/ai/nano-banana-2/}},
  note         = {Accessed: 2026-05-12}
}

@inproceedings{shi2025animaker,
  title={AniMaker: Multi-Agent Animated Storytelling with MCTS-Driven Clip Generation},
  author={Shi, Haoyuan and Li, Yunxin and Chen, Xinyu and Wang, Longyue and Hu, Baotian and Zhang, Min},
  booktitle={Proceedings of the SIGGRAPH Asia 2025 Conference Papers},
  pages={1--11},
  year={2025}
}

@article{wu2025automated,
  title={Automated movie generation via multi-agent cot planning},
  author={Wu, Weijia and Zhu, Zeyu and Shou, Mike Zheng},
  journal={arXiv preprint arXiv:2503.07314},
  year={2025}
}

@incollection{zhang2025anime,
  title={Anime: Adaptive multi-agent planning for long animation generation},
  author={Zhang, Lisai and Xu, Baohan and Yang, Siqian and Yin, Mingyu and Liu, Jing and Xu, Chao and Wang, Siqi and Wu, Yidi and Hong, Yuxin and Zhang, Zihao and others},
  booktitle={Proceedings of the SIGGRAPH Asia 2025 Posters},
  pages={1--3},
  year={2025}
}

@article{huang2026vimax,
  title={ViMax: Agentic Video Generation},
  author={Huang, Lingxuan and He, Sizhe and Zhou, Hengji and Nie, Liqiang and Xia, Lianghao and Huang, Chao},
  journal={arXiv preprint arXiv:2606.07649},
  year={2026}
}

@article{seedance2026seedance,
  title={Seedance 2.0: Advancing video generation for world complexity},
  author={Seedance, Team and Chen, De and Chen, Liyang and Chen, Xin and Chen, Ying and Chen, Zhuo and Chen, Zhuowei and Cheng, Feng and Cheng, Tianheng and Cheng, Yufeng and others},
  journal={arXiv preprint arXiv:2604.14148},
  year={2026}
}

@inproceedings{huang2024vbench,
  title={Vbench: Comprehensive benchmark suite for video generative models},
  author={Huang, Ziqi and He, Yinan and Yu, Jiashuo and Zhang, Fan and Si, Chenyang and Jiang, Yuming and Zhang, Yuanhan and Wu, Tianxing and Jin, Qingyang and Chanpaisit, Nattapol and others},
  booktitle={Proceedings of the IEEE/CVF Conference on Computer Vision and Pattern Recognition},
  pages={21807--21818},
  year={2024}
}

@inproceedings{wu2023human,
  title={Human preference score: Better aligning text-to-image models with human preference},
  author={Wu, Xiaoshi and Sun, Keqiang and Zhu, Feng and Zhao, Rui and Li, Hongsheng},
  booktitle={Proceedings of the IEEE/CVF International Conference on Computer Vision},
  pages={2096--2105},
  year={2023}
}

@article{wu2023q,
  title={Q-align: Teaching lmms for visual scoring via discrete text-defined levels},
  author={Wu, Haoning and Zhang, Zicheng and Zhang, Weixia and Chen, Chaofeng and Liao, Liang and Li, Chunyi and Gao, Yixuan and Wang, Annan and Zhang, Erli and Sun, Wenxiu and others},
  journal={arXiv preprint arXiv:2312.17090},
  year={2023}
}

@inproceedings{radford2021learning,
  title={Learning transferable visual models from natural language supervision},
  author={Radford, Alec and Kim, Jong Wook and Hallacy, Chris and Ramesh, Aditya and Goh, Gabriel and Agarwal, Sandhini and Sastry, Girish and Askell, Amanda and Mishkin, Pamela and Clark, Jack and others},
  booktitle={International conference on machine learning},
  pages={8748--8763},
  year={2021},
  organization={PmLR}
}

@inproceedings{tang2024dreamgaussian,
  title={Dreamgaussian: Generative gaussian splatting for efficient 3d content creation},
  author={Tang, Jiaxiang and Ren, Jiawei and Zhou, Hang and Liu, Ziwei and Zeng, Gang},
  booktitle={International Conference on Learning Representations},
  volume={2024},
  pages={33879--33896},
  year={2024}
}

@article{ren2023dreamgaussian4d,
  title={Dreamgaussian4d: Generative 4d gaussian splatting},
  author={Ren, Jiawei and Pan, Liang and Tang, Jiaxiang and Zhang, Chi and Cao, Ang and Zeng, Gang and Liu, Ziwei},
  journal={arXiv preprint arXiv:2312.17142},
  year={2023}
}

@article{yin20234dgen,
  title={4dgen: Grounded 4d content generation with spatial-temporal consistency},
  author={Yin, Yuyang and Xu, Dejia and Wang, Zhangyang and Zhao, Yao and Wei, Yunchao},
  journal={arXiv preprint arXiv:2312.17225},
  year={2023}
}

@article{liang2024diffusion4d,
  title={Diffusion4d: Fast spatial-temporal consistent 4d generation via video diffusion models},
  author={Liang, Hanwen and Yin, Yuyang and Xu, Dejia and Liang, Hanxue and Wang, Zhangyang and Plataniotis, Konstantinos N and Zhao, Yao and Wei, Yunchao},
  journal={arXiv preprint arXiv:2405.16645},
  year={2024}
}

@inproceedings{xing2025tip4gen,
  title={Tip4gen: Text to immersive panorama 4d scene generation},
  author={Xing, Ke and Liang, Hanwen and Xu, Dejia and Yin, Yuyang and Plataniotis, Konstantinos N and Zhao, Yao and Wei, Yunchao},
  booktitle={Proceedings of the 33rd ACM International Conference on Multimedia},
  pages={9267--9276},
  year={2025}
}

@article{yang2025matrix,
  title={Matrix-3d: Omnidirectional explorable 3d world generation},
  author={Yang, Zhongqi and Ge, Wenhang and Li, Yuqi and Chen, Jiaqi and Li, Haoyuan and An, Mengyin and Kang, Fei and Xue, Hua and Xu, Baixin and Yin, Yuyang and others},
  journal={arXiv preprint arXiv:2508.08086},
  year={2025}
}

@article{yin2025panoworld,
  title={Panoworld-x: Generating explorable panoramic worlds via sphere-aware video diffusion},
  author={Yin, Yuyang and Guo, HaoXiang and Liu, Fangfu and Wang, Mengyu and Liang, Hanwen and Li, Eric and Wang, Yikai and Jin, Xiaojie and Zhao, Yao and Wei, Yunchao},
  journal={arXiv preprint arXiv:2509.24997},
  year={2025}
}

@article{li2026martian,
  title={Martian world model: Controllable video synthesis with physically accurate 3d reconstructions},
  author={Li, Longfei and Fan, Zhiwen and Cong, Wenyan and Liu, Xinhang and Yin, Yuyang and Foutter, Matt and Pan, Panwang and You, Chenyu and Wang, Yue and Wang, Zhangyang and others},
  journal={Advances in Neural Information Processing Systems},
  volume={38},
  year={2026}
}

@article{team2026hy,
  title={HY-World 2.0: A Multi-Modal World Model for Reconstructing, Generating, and Simulating 3D Worlds},
  author={Team, HY and Cao, Chenjie and Zuo, Xuhui and Wang, Zhenwei and Zhang, Yisu and Wu, Junta and Liu, Zhenyang and Gong, Yuning and Liu, Yang and Yuan, Bo and others},
  journal={arXiv e-prints},
  pages={arXiv--2604},
  year={2026}
}

@misc{hitsztmg2026videoclaw,
  author={{HITsz-TMG}},
  title={VideoClaw: AI Creative Video Production Agent},
  howpublished={\url{https://github.com/HITsz-TMG/VideoClaw}},
  year={2026},
  note={GitHub repository, accessed 2026-08-08}
}

@misc{hbai2026toonflow,
  author={{HBAI-Ltd}},
  title={Toonflow-app: Open-source AI Tool for Animated Short Drama Creation},
  howpublished={\url{https://github.com/HBAI-Ltd/Toonflow-app}},
  year={2026},
  note={GitHub repository, accessed 2026-08-08}
}
}

\clearpage

\end{document}